\documentclass[letterpaper, 10 pt, conference]{ieeeconf}  

\IEEEoverridecommandlockouts                              

\usepackage[pdftex]{graphicx}
\usepackage{graphics} 
\usepackage{epsfig} 
\usepackage{mathptmx} 
\usepackage{times} 
\usepackage{amsmath} 
\usepackage{amssymb}  
\usepackage{algorithm}
\usepackage[noend]{algpseudocode}
\usepackage{multirow}
\usepackage{tikz}
\usepackage{lettrine}
\usepackage{color}
\usepackage{hyperref}

\newboolean{highlight}
\setboolean{highlight}{false}
\newcommand{\change}[1]{\ifthenelse{\boolean{highlight}}{\textcolor{red}{#1}}{#1}}

\newcommand{\secref}[1]{Section~\ref{#1}}

\newcommand{\figref}[1]{Fig.~\ref{#1}}
\newcommand{\myparagraph}[1]{\vspace{0.03in}\noindent\textbf{#1}}

\newcommand*\encircle[1]{\raisebox{0.6pt}{\textcircled{\raisebox{-0.2pt} {#1}}}}

\title{\LARGE \bf
Gel\textit{Slim}: A High-Resolution, Compact, Robust, \\ and Calibrated Tactile-sensing Finger 
\vspace{-4mm}
}

\author{Elliott Donlon, Siyuan Dong, Melody Liu, Jianhua Li, Edward Adelson and Alberto Rodriguez\\
Massachusetts Institute of Technology
\vspace{-4mm} \\ 
\thanks{Corresponding author: Elliott Donlon, {\tt\small <edonlon@mit.edu>}.} \thanks{This work was supported by the Karl Chang Innovation Fund, ABB, and the Toyota Research Institute.}
\thanks{The authors would like to thank Wenzhen Yuan for her expertise in making gels and willingness to teach; Francois Hogan, Maria Bauza, and Oleguer Canal for their use of the sensor and feedback along the way. Thanks also to Rachel Holladay for her totally awesome paper feedback.}}

\begin{document}

\maketitle
\thispagestyle{empty}
\pagestyle{empty}

\begin{abstract}
This work describes the development of a high-resolution tactile-sensing finger for robot grasping. This finger, inspired by previous GelSight sensing techniques, features an integration that is slimmer, more robust, and with more homogeneous output than previous vision-based tactile sensors. 
%

To achieve a compact integration, we redesign the optical path from illumination source to camera by combining light guides and an arrangement of mirror reflections.
We parameterize the optical path with geometric design variables and describe the tradeoffs between the finger thickness, the depth of field of the camera, and the size of the tactile sensing area.

The sensor sustains the wear from continuous use -- and abuse -- in grasping tasks by combining tougher materials for the compliant soft gel, a textured fabric skin, a structurally rigid body, and a calibration process that maintains homogeneous illumination and contrast of the tactile images during use.
Finally, we evaluate the sensor's durability along four metrics that track the signal quality during more than 3000 grasping experiments. 

\end{abstract}

\section{Introduction}








Tight integration of sensing hardware and control is key to 
mastery of manipulation in cluttered, occluded, or dynamic environments.
Artificial tactile sensors, however, are challenging to integrate and maintain: They are most useful when located at the distal end of the manipulation chain (where space is tight); they are subject to high-forces and wear (which reduces their life span or requires tedious maintenance procedures); and they require instrumentation capable of routing and processing high-bandwidth data.





Among the many tactile sensing technologies developed in the last decades~\cite{tactile_review_newer}, vision-based tactile sensors are a promising variant. They provide high spatial resolution with compact instrumentation and are synergistic with recent image-based deep learning techniques. 
Current implementations of these sensors, however, are often bulky and/or fragile~\cite{Tactilesensor_TacTip_edge,GelSightUSB,GelSight_Dong}. Robotic grasping benefits from sensors that are compactly-integrated and that are rugged enough to sustain the shear and normal forces involved in grasping.

\begin{figure}[t]
\centering
  \vspace{2mm}
  \includegraphics[width=\linewidth]{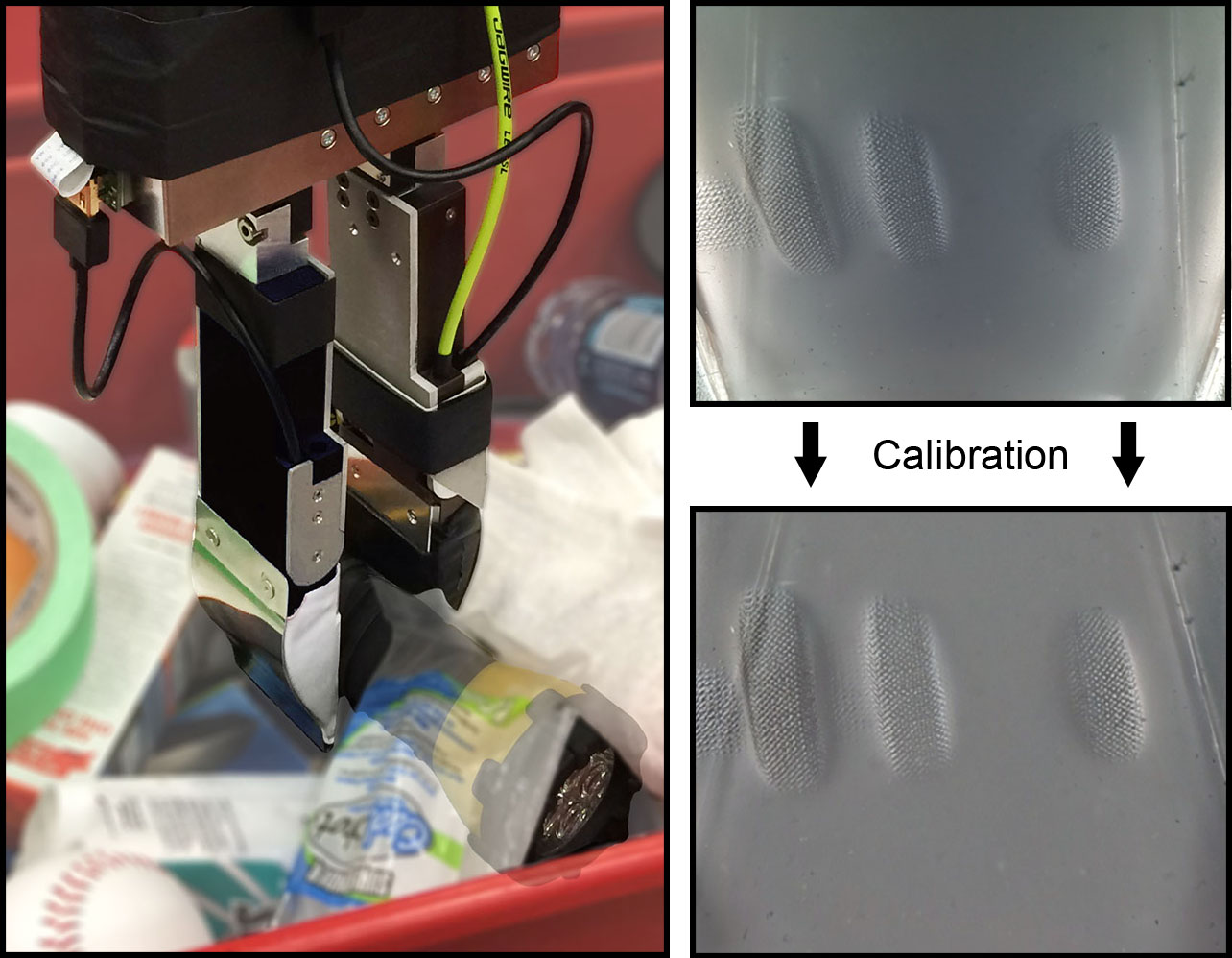}
  \caption{{\bf GelSlim fingers} picking a textured flashlight from clutter with the corresponding tactile image at right. The sensor is calibrated to normalize output over time and reduce the effects of wear on signal quality. The flashlight, though occluded, is shown for the reader's clarity.}
  \vspace{-2mm}
  \label{fig:teaser}
\end{figure}

\begin{figure*}[t]
\centering
  \vspace{2mm}
  \includegraphics[width=\textwidth]{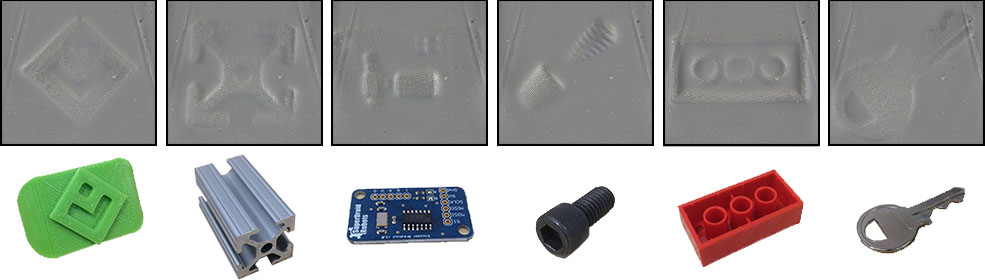}
  \caption{{\bf Tactile imprints.} From left to right: The MCube Lab's logo, 80/20 aluminum extrusion, a PCB, a screw, a Lego brick, and a key.}
  \vspace{-2mm}
  \label{fig:showing-off}
\end{figure*}

To address this need we present a tactile-sensing finger, \textit{GelSlim}, designed for grasping in cluttered environments (\figref{fig:teaser}). This finger, similar to other vision-based tactile sensors, uses a camera to measure tactile imprints (\figref{fig:showing-off}).\\

\begin{figure}[t]
\centering
  \vspace{2mm}
  \includegraphics[width=\linewidth]{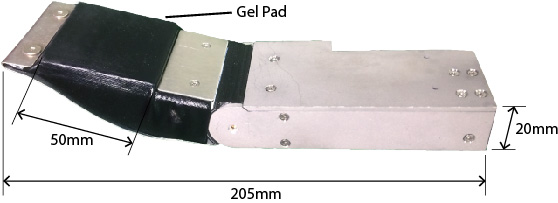}
  \caption{{\bf GelSlim finger.} Pointed adaptation of the GelSight sensor featuring a larger 50mm $\times$ 50mm sensor pad and strong, slim construction.}
  \vspace{-2mm}
  \label{fig:sensor}
\end{figure}

In this work we present: 
\begin{itemize}
\item \textbf{Design} of a vision-based high-resolution tactile-sensing finger with the form factor necessary to gain access to cluttered objects, and toughness to sustain the forces involved in everyday grasping (\secref{sec:design}). \change{The sensor outputs raw images of the sensed surface, which encode shape and texture of the object at contact.}
\item \textbf{Calibration} framework to regularize the sensor output over time and across sensor individuals (\secref{sec:calibration}). We suggest four metrics to track the quality of the tactile feedback.
\item \textbf{Evaluation} of the sensor's durability by monitoring its image quality over more than 3000 grasps (\secref{sec:calibration}).
\end{itemize}

The long term goal of this research is to enable reactive grasping and manipulation. 
The use of tactile feedback in the control loop of robotic manipulation is key for reliability. 
Our motivation stems from efforts in developing bin-picking systems to grasp novel objects in cluttered scenes and from the need to observe the geometry of contact to evaluate and control the quality of a grasp~\cite{Zeng2017, Zeng2018}.

In cluttered environments like a pantry or a warehouse storage cell, as in the Amazon Robotics Challenge~\cite{Correll2016}, a robot faces the challenge of singulating target objects from a tightly-packed collection of items. 
Cramped spaces and clutter lead to frequent contact with non-target objects. Fingers must be compact and, when possible, pointed to squeeze between target and clutter (\figref{fig:sensor}). To make use of learning approaches, tactile sensors must also be resilient to the wear and tear from long experimental sessions which often yield unexpected collisions. Finally, sensor calibration and signal conditioning are key to the consistency of tactile feedback as the sensor's physical components decay.

%

\section{Related Work}



The body of literature on tactile sensing technologies is large~\cite{tactile_review_newer,Tactilesensor_review1}. Here we discuss relevant works related to the technologies used by the proposed sensor: vision-based tactile sensors and GelSight sensors.

\subsection{Vision-based tactile sensors}

Cameras provide high-spatial-resolution 2D signals without the need for many wires. Their sensing field and working distance can also be tuned with an optical lens. For these reasons, cameras are an interesting alternative to several other sensing technologies, which tend to have higher temporal bandwidth but more limited spatial resolution.

Ohka~\textit{et al}.~\cite{Tactilesensor_array1996} designed an early vision-based tactile sensor. It is comprised of a flat rubber sheet, an acrylic plate and a CCD camera to measure three-dimensional force. The prototyped sensor, however, was too large to be realistically integrated in a practical end-effector. GelForce~\cite{Tactilesensor_Gelforce}, a tactile sensor shaped like a human finger, used a camera to track two layers of dots on the sensor surface to measure both the magnitude and direction
of an applied force. 

Instead of measuring force, some vision-based tactile sensors focus on measuring geometry, such as edges, texture or 3D shape of the contact surface. Ferrier and Brockett~\cite{Tactilesensor_dotshape} proposed an algorithm to reconstruct the 3D surface by analyzing the distribution of the deformation of a set of markers on a tactile sensor. This principle has inspired several other contributions. The TacTip sensor~\cite{Tactilesensor_TacTip_edge} uses a similar principle to detect edges and estimate the rough 3D geometry of the contact surface. Mounted on a GR2 gripper, the sensor gave helpful feedback when reorienting a cylindrical object in hand~\cite{Tactilesensor_TacTip_inhand}.
Yamaguchi~\cite{Tactilesensor_CMU} built a tactile sensor with a clear silicone gel that can be mounted on a Baxter hand. Unlike the previous sensors, Yamaguchi's also captures the local color and shape information since the sensing region is transparent. The sensor was used to detect slip and estimate contact force.


\subsection{GelSight sensors}

The GelSight sensor is a vision-based tactile sensor that measures the 2D texture and 3D topography of the contact surface. It utilizes a piece of elastomeric gel with an opaque coating as the sensing surface, and a webcam above the gel to capture contact deformation from changes in lighting contrast as reflected by the opaque coating. 
The gel is illuminated by color LEDs with inclined angles and different directions. The resulting colored shading can be used to reconstruct the 3D geometry of the gel deformation. The original, larger GelSight sensor~\cite{GelSight2009,GelSight2011} was designed to measure the 3D topography of the contact surface with micrometer-level spatial resolution. Li~\textit{et al}.~\cite{GelSightUSB} designed a cuboid fingertip version that could be integrated in a robot finger. Li's sensor has a $1\times1$ cm$^2$ sensing area, and can measure fine 2D texture and coarse 3D information. A new version of the GelSight sensor was more recently proposed by Dong~\textit{et al}.~\cite{GelSight_Dong} to improve 3D geometry measurements and standardize the fabrication process. A detailed review of different versions of GelSight sensors can be found in ~\cite{GelSight_review}. 

GelSight-like sensors with rich 2D and 3D information have been successfully applied in robotic manipulation. 
Li~\textit{et al}.~\cite{GelSightUSB} used GelSight's localization capabilities to insert a USB connector, where the sensor used the texture of the characteristic USB logo to guide the insertion. 
Izatt~\textit{et al}.~\cite{GelSightRuss} explored the use of the 3D point cloud measured by a GelSight sensor in a state estimation filter to find the pose of a grasped object in a peg-in-hole task.
%
Dong~\textit{et al}.~\cite{GelSight_Dong} used the GelSight sensor to detect slip from variations in the 2D texture of the contact surface in a robot picking task. The 2D image structure of the output from a GelSight sensor makes it a good fit for deep learning architectures. GelSight sensors have also been used to estimate grasp quality~\cite{calandra2017feeling}.

\subsection{Durability of tactile sensors}

Frictional wear is an issue intrinsic to tactile sensors. Contact forces and torques during manipulation are significant and can be harmful to both the sensor surface and its inner structure. 
Vision-based tactile sensors are especially sensitive to frictional wear, since they rely on the deformation of a soft surface for their sensitivity. These sensors commonly use some form of soft silicone gel, rubber or other soft material as a sensing surface~\cite{Tactilesensor_Gelforce,Tactilesensor_TacTip,Tactilesensor_CMU, GelSight_Dong}.

To enhance the durability of the sensor surface, researchers have investigated using protective skins such as plastic~\cite{Tactilesensor_CMU}, or making the sensing layer easier to replace by involving 3D printing techniques with soft material~\cite{Tactilesensor_TacTip}.

Another mechanical weakness of vision-based tactile sensors is the adhesion between the soft sensing layer and its stronger supporting layer. Most sensors discussed above use either silicone tape or rely on the adhesive property of the silicone rubber, which can be insufficient under practical shear forces involved in picking and lifting objects. The wear effects on these sensors are especially relevant if one attempts to use them in a data-driven/learning context~\cite{Tactilesensor_CMU,calandra2017feeling}.

Durability is key to the practicality of a tactile sensor; however, none of the above provide quantitative analysis of their sensor's durability over usage. 

\section{Design Goals}
\label{sec:goals}

\begin{figure}[t]
\centering
  \vspace{2mm}
  \includegraphics[width=6cm]{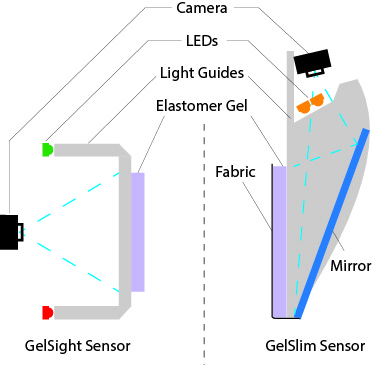}
  \caption{{\bf The construction of a GelSight sensor.} A general integration of a GelSight sensor in a robot finger requires three components: camera, light, and gel, in particular arrangement. Li's original fingertip schematic \cite{GelSightUSB} is shown at left with our \textit{GelSlim} at right.}
  \vspace{-2mm}
  \label{fig:GelSight-construction}
\end{figure}

In a typical GelSight-like sensor, a clear gel with an opaque outer membrane is illuminated by a light source and captured by a camera (\figref{fig:GelSight-construction}). The position of each of these elements depends on the specific requirements of the sensor. Typically, for ease of manufacturing and optical simplicity, the camera's optical axis is normal to the gel (left of \figref{fig:GelSight-construction}).
To reproduce 3D using photometric techniques~\cite{GelSight2011}, at least three colors of light must be directed across the gel from different directions. 

Both of these geometric constraints, the camera placement and the illumination path, are counterproductive to slim robot finger integrations, and existing sensor implementations are cuboid. In most manipulation applications, successful grasping requires fingers with the following qualities:

\begin{itemize}
\item \textbf{Compactness} allows fingers to singulate objects from clutter by squeezing between them or separating them from the environment.
\item \textbf{Uniform Illumination} makes sensor output consistent across as much of the gel pad as possible. 
\item \textbf{Large Sensor Area} extends the area of the tactile cues, both where there is and where there is not contact. This can provide a better knowledge of the state of the grasped object and, ultimately, enhanced controllability.
\item \textbf{Durability} affords signal stability, necessary for the time-span of the sensor. This is especially important for data-driven techniques that build models from experience.
\end{itemize}

In this paper we propose a redesign of the form, materials, and processing of the GelSight sensor to turn it into a GelSight \emph{finger}, yielding a more useful finger shape with a more consistent and calibrated output (right of \figref{fig:GelSight-construction}).
The following sections describe the geometric and optical tradeoffs in its design (\secref{sec:design}), as well as the process to calibrate and evaluate it (\secref{sec:calibration}).

\section{Design and Fabrication}
\label{sec:design}


\begin{figure}[b]
\centering
  \vspace{2mm}
  \includegraphics[width=\linewidth]{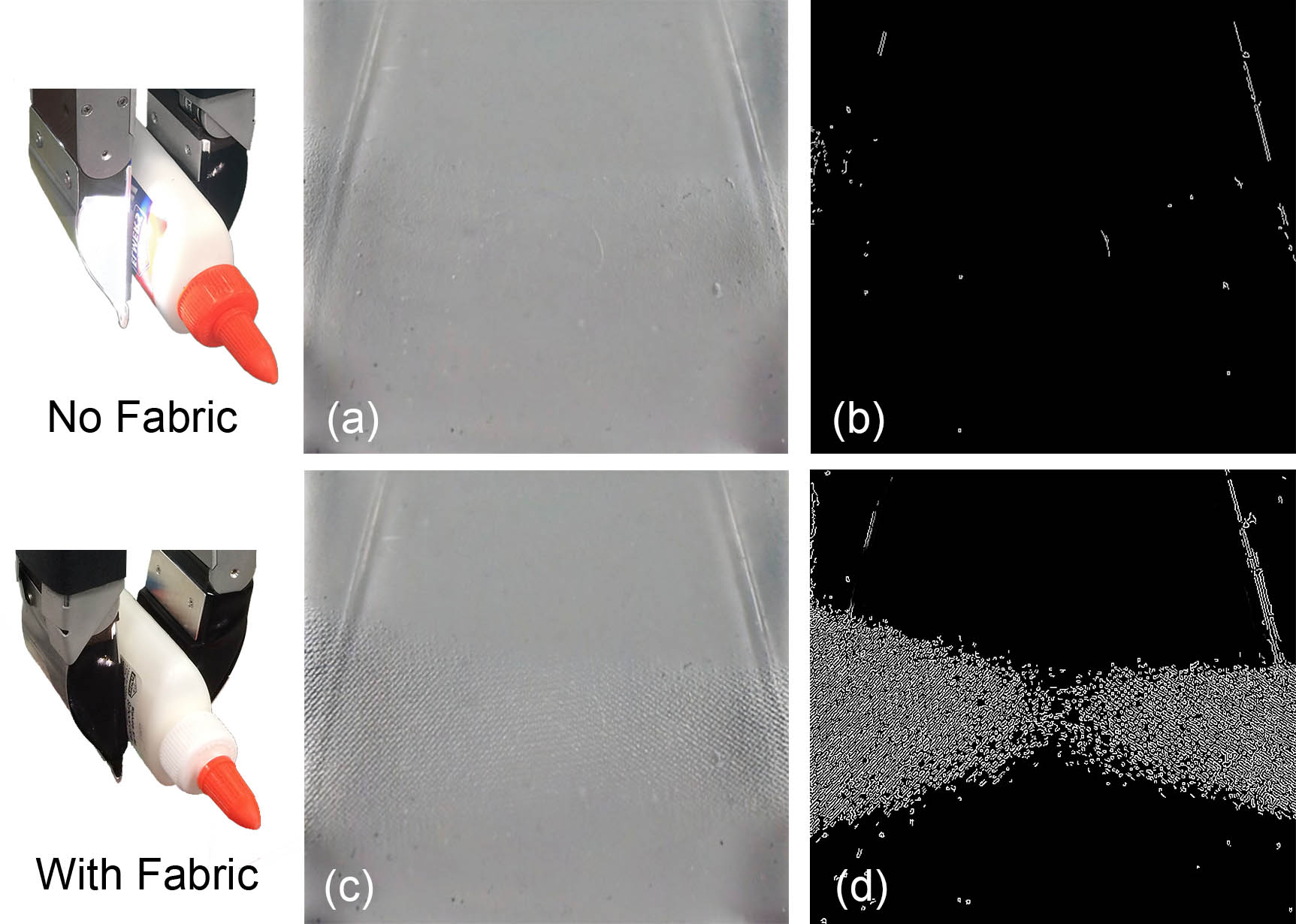}
  \caption{{\bf Texture in the sensor fabric skin improves signal strength.} When an object with no texture is grasped against the gel with no fabric, signal is very low (a-b). The signal improves with textured fabric skin (c-d). The difference stands out well when processed with Canny edge detection.}
  \vspace{-2mm}
  \label{fig:covered-difference}
\end{figure}


To realize the design goals in \secref{sec:goals}, we propose the following changes to a standard design of a vision-based GelSight-like sensor:
%
%
1) Photometric stereo for 3D reconstruction requires precise illumination. Instead, we focus on recovering texture and contact surface, which will allow more compact light-camera arrangements. 
2) The softness of the gel plays a role in the resolution of the sensor, but is also damaging to its life span. We will achieve higher durability by protecting the softest component of the finger, the gel, with textured fabric. 
3) Finally, we will improve the finger's compactness, illumination uniformity, and sensor pad size with a complete redesign of the sensor optics.

\subsection{Gel Materials Selection}

A GelSight sensor's gel must be elastomeric, optically clear, soft, and resilient. Gel hardness represents a tradeoff between spatial resolution and strength. Maximum sensitivity and resolution is only possible when gels are very soft, but their softness yields two major drawbacks: low tensile strength and greater viscoelasticity.
Given our application's lesser need for spatial resolution, we make use of slightly harder, more resilient gels compared to other Gelsight sensors~\cite{GelSightUSB,GelSight_Dong}. Our final gel formulation is a two-part silicone (XP-565 from Silicones Inc.) mixed in a 15:1 ratio of parts A to B. The outer surface of our gel is coated with a specular silicone paint using a process developed by Yuan~\textit{et al}.~\cite{GelSight_review}. 

The surface is covered with a stretchy, loose-weave fabric to prevent damage to the gel while increasing signal strength. Signal strength is proportional to deformation due to pressure on gel surface. Because the patterned texture of the fabric lowers the contact area between object and gel, pressure is increased to the point where the sensor can detect the contact patch of flat objects pressed against the flat gel (\figref{fig:covered-difference}).

\subsection{Sensor Geometry Design Space}

\begin{figure}[t]
\centering
  \includegraphics[width=8cm]{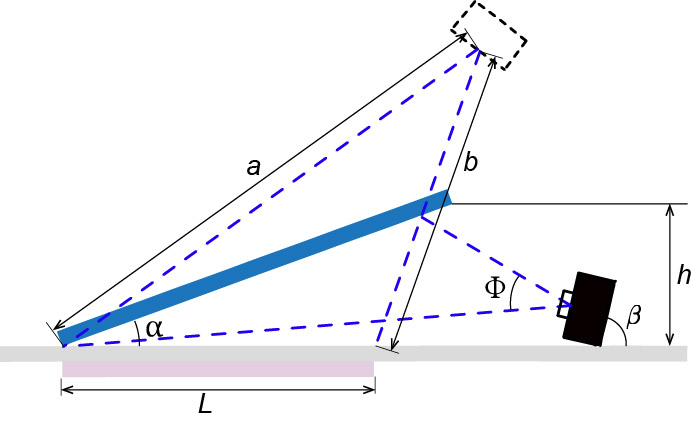}
  \caption{{\bf The design space of a single-reflection GelSight sensor.} Based on the camera's depth of field and viewing angle, it will lie at some distance away from the gel. These parameters, along with mirror and camera angles, determine the thickness of the finger and the size of the gel pad. The virtual camera created by the mirror is drawn for visualization purposes.}
  \label{fig:parametric-wedge}
  \vspace{-3mm}
\end{figure}

We change the sensor's form factor by using a mirror to reflect the gel image back to the camera. This allows us to have a larger sensor pad by placing the camera farther away while also keeping the finger comparatively thin. A major component of finger thickness is the optical region with thickness $h$ shown in \figref{fig:parametric-wedge}, which is given by the trigonometric relation:

\begin{equation} \label{equ1}
h = \frac{L \cdot \cos(\beta-\frac{\Phi}{2}-2\alpha)}{\cos(\frac{\Phi}{2}-\beta+\alpha)}\,,
\end{equation}
%
where $\Phi$ is the camera's field of view, $\alpha$ is mirror angle, $\beta$ is the camera angle relative to the base, and $L$ is the length of the gel. $L$ is given by the following equation and also relies on the disparity between the shortest and longest light path from the camera (depth of field):
\begin{equation} \label{equ4}
L = \frac{(a-b) \cdot \sin{\Phi}}{2\sin{\frac{\Phi}{2}} \cdot \sin{(\beta-2\alpha)}}\,.
\end{equation}
%
%

Together, the design requirements $h$ and $L$, vary with the design variables, $\alpha$ and $\beta$, and are constrained by the camera's depth of field: $(a-b)$ and viewing angle: $\Phi$. These design constraints ensure that both near and far edges given by \eqref{equ4} are in focus and that the gel is maximally sized and the finger is minimally thick.

\subsection{Optical Path: Photons From Source, to Gel, to Camera}

\begin{figure}[t]
\centering
  \includegraphics[width=7.5cm]{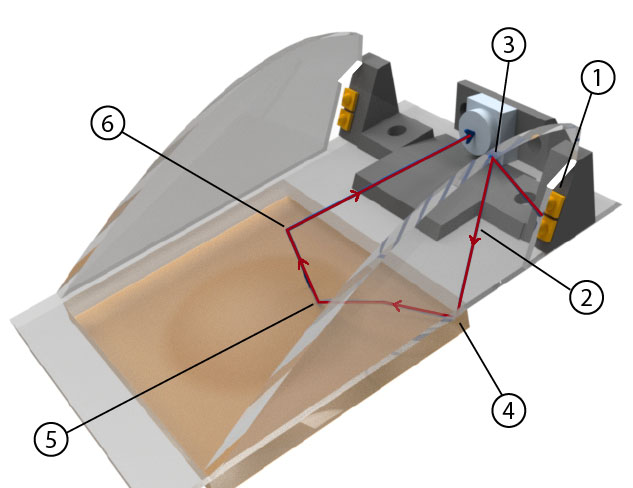}
  \caption{{\bf The journey of a light ray through the finger.} The red line denoting the light ray is: 1) Emitted by two compact, high-powered LEDs on each side. 2) Routed internal to acrylic guides via total internal reflection. 3) Redistributed to be parallel and bounced toward the gel pad by a parabolic mirror. 4) Reflected $90^{\circ}$ on a mirror surface to graze across the gel. 5) Reflected up by an object touching the gel. 6) Reflected to the camera by a flat mirror (not shown in the figure).}
  \label{fig:light-bouncing}
  \vspace{-3mm}
\end{figure}

Our method of illuminating the gel makes three major improvements relative to previous sensors: a slimmer finger tip, more even illumination, and a larger gel pad. Much like Li did in his original GelSight finger design~\cite{GelSightUSB}, we use acrylic wave guides to move light throughout the sensor with fine control over the angle of incidence across the gel (\figref{fig:light-bouncing}). However, our design moves the LEDs used for illumination farther back in the finger by using an additional reflection, thus allowing our finger to be slimmer at the tip. 

The light cast on the gel originates from a pair of high-powered, neutral white, surface-mount LEDs (OSLON SSL 80) on each side of the finger. Light rays stay inside the acrylic wave guide due to total internal reflection by the difference in refractive index between acrylic and air. 
Optimally, light rays would be emitted parallel so as to not lose intensity as light is cast across the gel. However, light emitters are usually point sources. A line of LEDs, as in Li's sensor, helps to evenly distribute illumination across one dimension while intensity decays across the length of the sensor. 

Our approach uses a parabolic reflection (Step 3 in \figref{fig:light-bouncing}) to ensure that light rays entering the gel pad are close to parallel. The two small LEDs are approximated as a single point source and placed at the parabola's focus. Parallel light rays bounce across the gel via a hard $90^{\circ}$ reflection. Hard reflections through acrylic wave guides are accomplished by painting those surfaces with mirror finish paint.

When an object makes contact with the fabric over the gel pad, it creates a pattern of light and dark spots as the specular gel interacts with the grazing light. This image of light and dark spots is transmitted back to the camera off a front-surface glass mirror. The camera (Raspberry Pi Spy Camera) was chosen for its small size, low price, high framerate/resolution, and good depth of field.

\subsection{Lessons Learned}

For robotic system integrators, or those interested in designing their own GelSight sensors, the following is a collection of small but important lessons we learned:
\begin{enumerate}
    \item{\bf{Mirror:}} Back surface mirrors create a ``double image" from reflections off front and back surfaces especially at the reflection angles we use in our sensor. Glass front surface mirrors give a sharper image.
    \item{\bf{Clean acrylic:}} Even finger oils on the surface of a wave guide can interrupt total internal reflection. Clean acrylic obtains maximum illumination efficiency.
    \item{\bf{Laser cut acrylic:}} Acrylic pieces cut by laser exhibit stress cracking at edges after contacting solvents from glue or mirror paint. Cracks break the optical continuity in the substrate and ruin the guide. Stresses can be relieved by annealing first.
    \item{\bf{LED choice:}} This LED was chosen for its high luminous efficacy (103 lm/W), compactness (3mm $\times$ 3mm), and small viewing angle (80$^\circ$). Small viewing angle directs more light into the thin wave guide.
    \item{\bf{Gel paint type:}} From our experience in this configuration, semi-specular gel coating provides a higher-contrast signal than lambertian gel coatings. Yuan~\textit{et al}. \cite{GelSight_review} describe the different types of coatings and how to manufacture them.
    \item{\bf{Affixing silicone gel:}} When affixing the silicone gel to the substrate, most adhesives we tried made the images hazy or did not acceptably adhere to either the silicone or substrate. We found that \textit{Adhesives Research ARclear 93495} works well. Our gel-substrate bond is also stronger than other gel-based sensors because of its comparatively large contact area.
\end{enumerate}
Some integration lessons revolve around the use of a Raspberry Pi spy camera. It enables a very high data-rate but requires a 15-pin Camera Serial Interface (CSI) connection with the Raspberry Pi. Since the GelSlim sensor was designed for use on a robotic system where movement and contact are part of normal operation, the processor (Raspberry Pi) is placed away from the robot manipulator. We extended the camera's fragile ribbon cable by first adapting it to an HDMI cable inside the finger, then passing that HDMI cable along the kinematic chain of the robot. Extending the camera this way allows us to make it up to several meters long, mechanically and electrically protect the contacts, and route power to the LEDs through the same cable.

The final integration of the sensor in our robot finger also features a rotating joint to change the angle of the finger tip relative to the rest of the finger body. This movement does not affect the optical system and allows us to more effectively grasp a variety of objects in clutter. 

\change{There are numerous ways to continue improving the sensor's durability and simplify the sensor's fabrication process. For example, while the finger is \textit{slimmer}, it is not \textit{smaller}. It will be a challenge to make integrations sized for smaller robots due to camera field of view and depth of field constraints. Additionally, our finger has an un-sensed, rigid tip that is less than ideal for two reasons: it is the part of the finger with the richest contact information, and its rigidity negatively impacts the sensor's durability. To decrease contact forces applied due to this rigidity, we will add compliance to the finger-sensor system.}

\subsection{Gel Durability Failures}

We experimented with several ways to protect the gel surface before selecting a fabric skin. Most non-silicone coatings will not stick to the silicone bulk, so we tested various types of filled and non-filled silicones. Because this skin coats the outside, using filled (tougher, non-transparent) silicones is an option. One thing to note is that thickness added outside of the specular paint increases impedance of the gel, thus decreasing resolution. To deposit a thin layer onto the bulk, we diluted filled, flowable silicone adhesive with NOVOCS silicone thinner from Smooth-On Inc. We found that using solvent in proportions greater than 2:1 (solvent:silicone) caused the gel to wrinkle (possibly because solvent diffused into the gel and caused expansion). 

Using a non-solvent approach to deposit thin layers like spin coating is promising, but we did not explore this path. Furthermore, thin silicone coatings often rubbed off after a few hundred grasps signaling that they did not adhere to the gel surface effectively. Plasma pre-treatment of the silicone surface could more effectively bond substrate and coating, but we were unable to explore this route.

\section{Sensor Calibration}
\label{sec:calibration}

\begin{figure}[t]
\centering
  \includegraphics[scale=0.55]{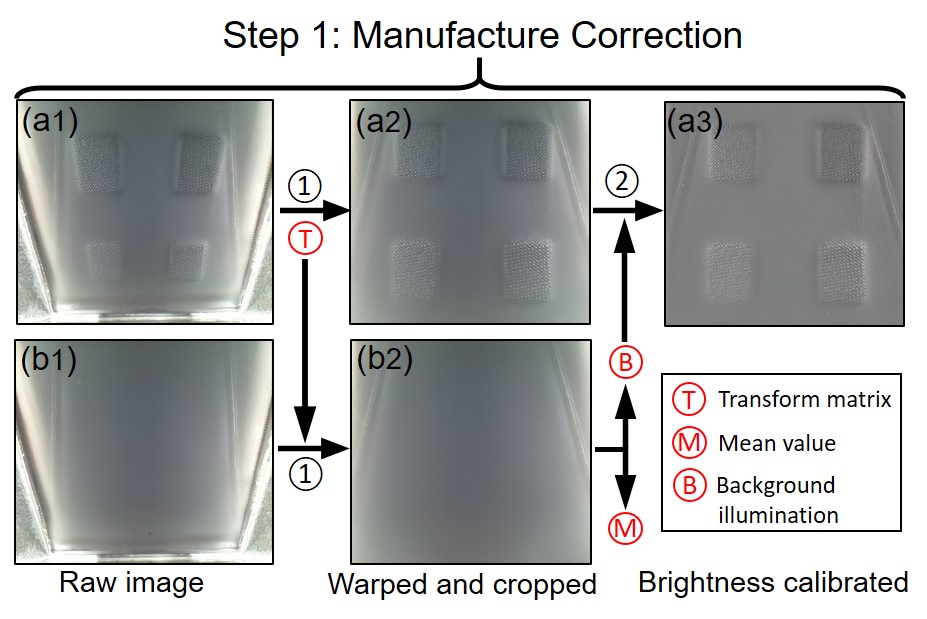}
  \caption{\textbf{Calibration Step 1 (manufacture correction)}: capture raw image (a1) against  a  calibration pattern with four rectangles and a non-contact image (b1); calculate the ``transform matrix" \encircle{{\tiny T}} according to image (a1); do operation \encircle{{\scriptsize 1}} ``image warping and cropping'' to image (a1) and (b1) and get (a2) and (b2); apply Gaussian filter to (b2) to get ``background illumination'' \encircle{{\tiny B}}; do operation \encircle{{\scriptsize 2}} to (a2) and get the calibrated image (a3); record the ``mean value'' \encircle{{\tiny M}} of image (b2) as brightness reference.}
  \label{fig:step1}
  \vspace{-3mm}
\end{figure}
\begin{figure}[t]
\centering
  \includegraphics[scale=0.55]{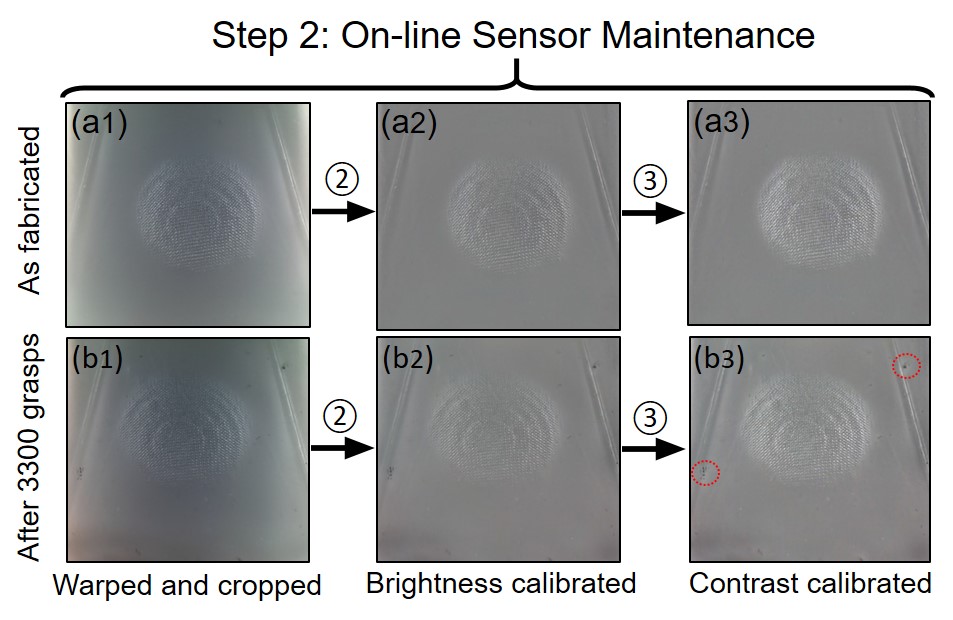}
  \caption{\textbf{Calibration Step 2 (on-line sensor maintenance)}: apply transformation \encircle{{\tiny T}} to start from a warped and cropped image (a1) and (b1); operation 
  \encircle{{\scriptsize 2}} uses the non-contact image from step 1 and adds constant \encircle{{\tiny M}} to calibrate the image brightness (a2) and (b2); operation \encircle{{\scriptsize 3}} performs a 
  local contrast adjustment (a3) and (b3). All the images show the imprint of the calibration ``dome" 
  after fabrication and 3300 grasps. \textcolor{black}{The red circles in b(3) highlight the region where the gel wears out after 3300 grasps.}}
  \label{fig:step2}
  \vspace{-3mm}
\end{figure}

The consistency of sensor output is key for sensor usability.
The raw image from a GelSlim sensor right after fabrication has two intrinsic issues: non-uniform illumination and a strong perspective distortion.
%
In addition, the sensor image stream may change during use due to small deformations of the hardware, compression of the gel, or camera shutter speed fluctuations.
%
%
To improve the consistency of the signal we introduce a two-step calibration process, illustrated in \figref{fig:step1} and~\figref{fig:step2}.

\begin{figure}[h]
\centering
  \includegraphics[width=\linewidth]{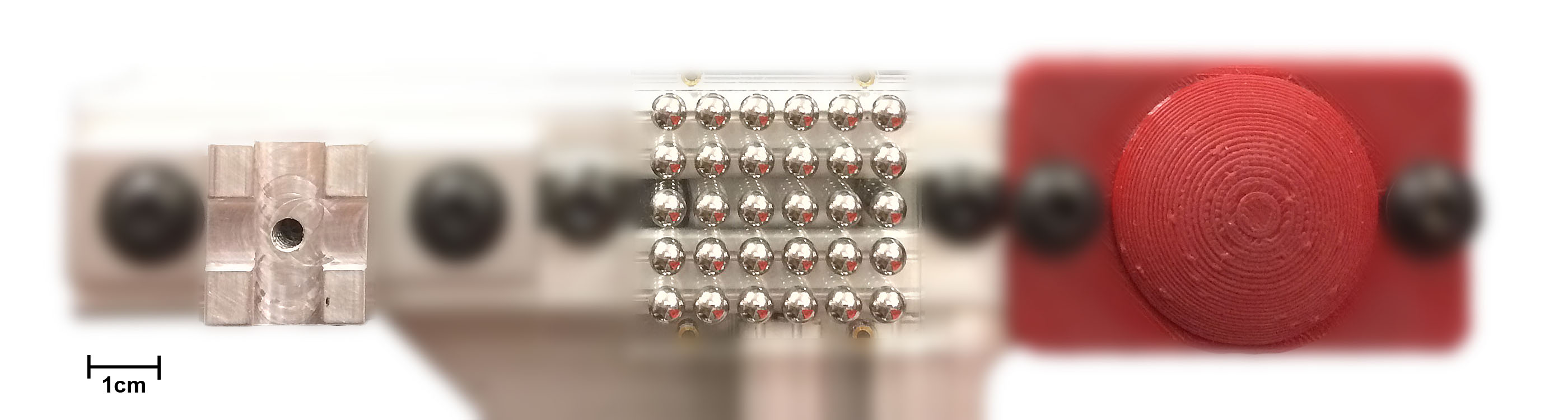}
  \caption{Three tactile profiles to calibrate the sensor. From left to right: A rectangle with sharp corners, a ball-bearing array, and a 3D printed dome.}
  \label{fig:calibration-targets}
  \vspace{-3mm}
\end{figure}

\myparagraph{Calibration Step 1. Manufacture correction.}
After fabrication, the sensor signal can vary with differences in camera perspective and illumination intensity.
To correct for camera perspective, we capture a tactile imprint in \figref{fig:step1} (a1) against a calibration pattern with four flat squares (\figref{fig:calibration-targets} left). 
%
With the distance between the outer edges of the four squares, we estimate the perspective transformation matrix $T$ that allows us to warp the image to a normal perspective and crop the boundaries.
The contact surface information in the warped image (\figref{fig:step1} (a2)) is more user-friendly. We assume the perspective camera matrix remains constant, so the manufacture calibration is done only once.

We correct for non-homogenous illumination by estimating the illumination distribution of the background $B$ using a strong Gaussian filter on a non-contact warped image (\figref{fig:step1} (b2)). The resulting image, after subtracting the non-uniform illumination background (\figref{fig:step1} (a3)), is visually more homogeneous. In addition, we record the mean value of the warped non-contact image $M$ as brightness reference for future use.

\myparagraph{Calibration Step 2. On-line Sensor Maintenance.}
The aim of the second calibration step is to keep the sensor output consistent over time. 
We define four metrics to evaluate the temporal consistency of the 2D signal: \textit{Light intensity and distribution}, \textit{Signal strength}, \textit{Signal strength distribution} and \textit{Gel condition}. In the following subsection, we will describe and evaluate these metrics in detail.

We will make use of the calibration targets in \figref{fig:calibration-targets} to track the signal quality, including a ball-bearing array and a 3D printed dome. We conduct over 3300 aggressive grasp-lift-vibrate experiments on several daily objects with two GelSlim fingers on a WSG-50 gripper \change{attached to an ABB IRB 1600ID robotic arm}. We take a tactile imprint of the two calibration targets every 100 grasps. \change{The data presented in the following sections were gathered with a single prototype and are for the purposes of evaluating sensor durability trends.}


\subsection{Metric I: Light Intensity and Distribution}

\begin{figure}[t]
\centering
  \includegraphics[width=\linewidth]{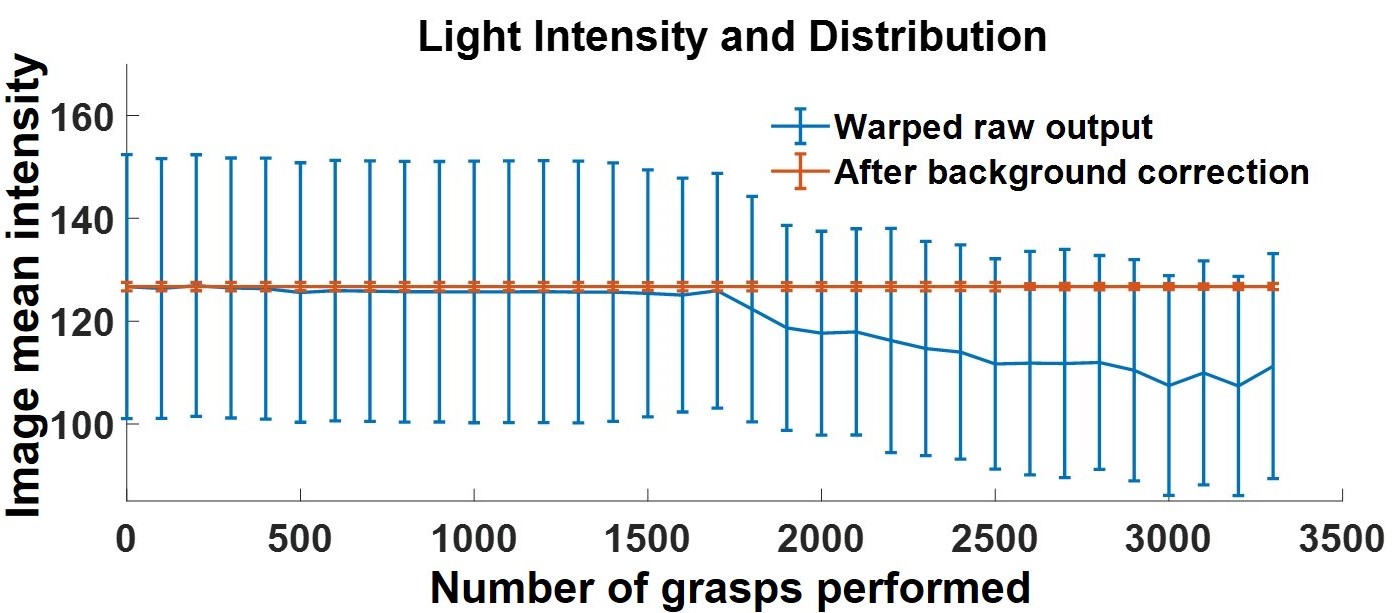}
  \caption{Evolution of the light intensity distribution.}
  \label{fig:light}
  \vspace{-3mm}
\end{figure}

The light intensity and distribution are the mean and standard deviation of the light intensity in a non-contact image.
The light intensity distribution in the gel is influenced by the condition of the light source, the consistency of the optical path and the homogeneity of the paint of the gel. The three factors can change due to wear.
%
%
\figref{fig:light} shows their evolution over grasps before (blue) and after (red) background illumination correction. The standard deviations are shown as error bars. 
%
%
The blue curve (raw output from the sensor) shows that the mean brightness of the image drops slowly over time, especially after around 1750 grasps.
This is likely due to slight damage of the optical path. The variation of the image brightness over space decreases slightly, which is likely caused by the fact that the bright two sides of the image get darker and more similar to the center region. \figref{fig:step2} shows an example of the decrease in illumination before (a1) and after (b1) 3300 grasps. 

We compensate for the changes in light intensity by subtracting the background and adding a constant $M$ (brightness reference from step one) to the whole image. The background illumination is obtained from the Gaussian filtered non-contact image at that point. The mean and variance of the corrected images, shown in red in \figref{fig:light}, are more consistent. \figref{fig:step2} shows an example of the improvement after 3300 grasps.

\subsection{Metric II: Signal Strength}

The signal strength $S$ is a measure of the dynamic range of the tactile image under contact.
It is intuitively the brightness and contrast of a contact patch, and we define it as:
\begin{equation} \label{equ3}
S = H(\sigma-m)(\frac{2\mu}{255}  + \frac{\sigma}{n})\,,
\end{equation}
where $\mu$ is the mean and $\sigma$ the standard deviation of the image intensity in the contact region, and $H(x)$ is the Heaviside step function. $H(\sigma-m)$ means that if the standard deviation is smaller than $m$, signal strength is 0. Experimentally, we set $m$ to 5, and $n$, the standard deviation normalizer, to 30.

Maintaining a consistent signal strength during use is one of the most important factors for the type of contact information we can extract in a vision-based tactile sensor. 
In a GelSlim sensor, signal strength is affected by the elasticity of the gel, which degrades after use.

\begin{figure}[b]
\centering
  \includegraphics[width=\linewidth]{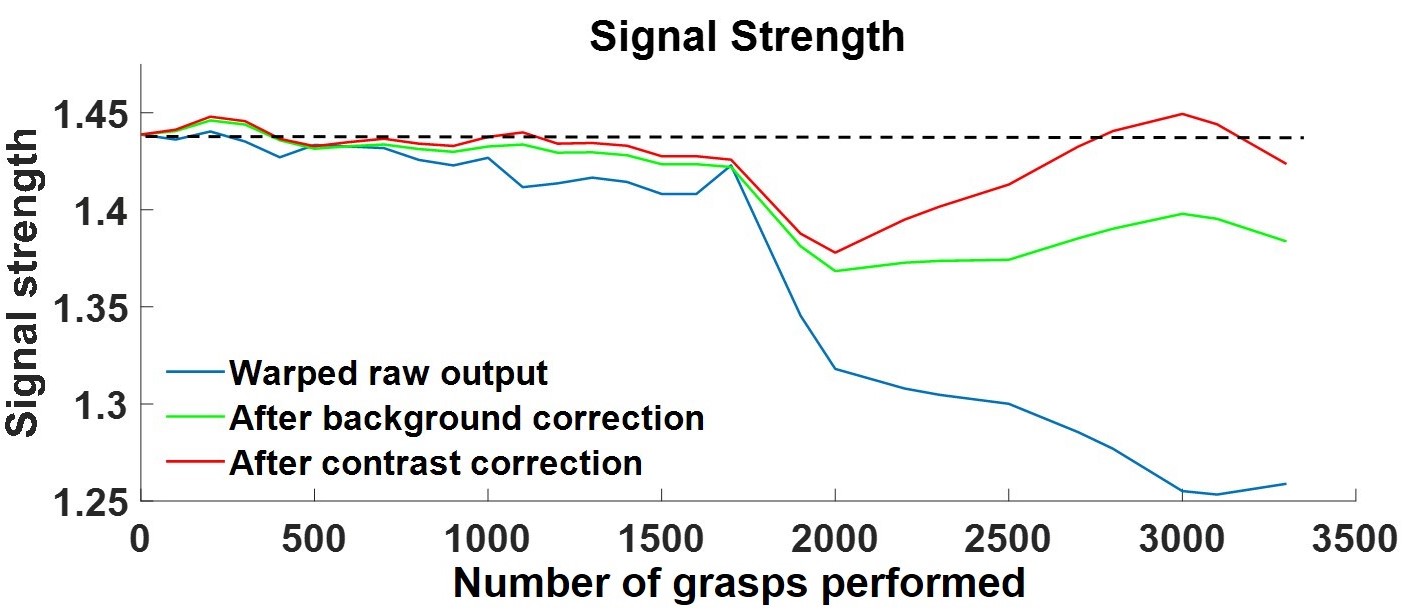}
  \caption{The change of signal strength across the number of grasps performed.}
  \label{fig:signal_strength}
  \vspace{-3mm}
\end{figure}

We track the signal strength during grasps by using the ``dome" calibration pattern designed to yield a single contact patch. \figref{fig:signal_strength} shows its evolution. The blue curve (from raw output) shows a distinct drop of the signal strength after 1750 grasps. \change{The brightness decrease described in the previous subsection is one of the key reasons}.  

The signal strength can be enhanced by increasing both the contrast and brightness of the image. The brightness adjustment done after fabrication improves the signal strength, shown in green in \figref{fig:signal_strength}. However, the image with brightness correction after 3300 grasps shown in \figref{fig:step2} (b2) still has decreased contrast.

To enhance the image contrast according to the illumination, we perform adaptive histogram equalization to the image, which increases the contrast on the whole image, and then fuses the images with and without histogram equalization together according to the local background illumination. The two images after the whole calibration are shown in \figref{fig:step2} (a3) and (b3). The signal strength after calibrating illumination and contrast (\figref{fig:signal_strength} in red) shows better consistency during usage.  

\subsection{Metric III: Signal Strength Distribution}

\begin{figure}[t]
\centering
  \includegraphics[width=\linewidth]{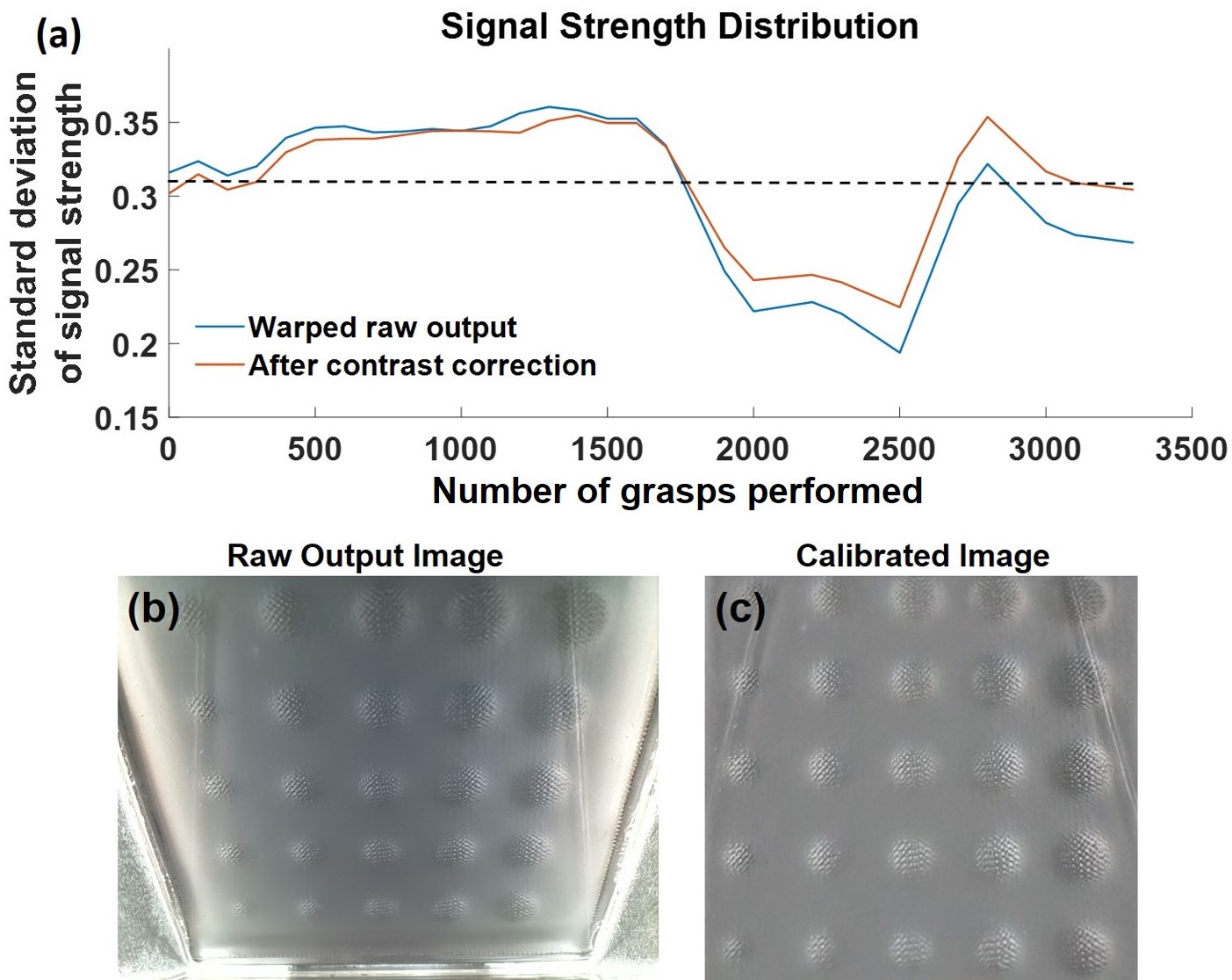}
  \caption{(a) The evolution of signal strength distribution (b) The raw output of ``ball array" calibration image (c) The calibrated ``ball array" calibration image.}
  \label{fig:SS_distribution}
  \vspace{-3mm}
\end{figure}

The force distribution after grasping an object is non-uniform across the gel.
During regular use, the center and distal regions of the gel are more likely to be contacted during grasping, which puts more wear on the gel in those areas. This phenomenon results in a non-uniform degradation of the signal strength.
%
%
To quantify this phenomenon, we extract the signal strength of each pressed region from the ``ball array" calibration images taken every 100 grasps (see \figref{fig:SS_distribution} (b) before and (c) after calibration). We use the standard deviation of the $5 \times 5$ array of signal strengths to represent the signal strength distribution, and compensate for variations by increasing the contrast non-uniformly in the decreased regions.

\figref{fig:SS_distribution} shows signal strength distribution before and after calibration in (blue) and (red) respectively. The red curve shows some marginal improvement in the consistency over usage. \change{The sudden increase of the curve after 2500 grasps is caused by the change in light distribution likely due to damage of the optical path by an especially aggressive grasp.}

\subsection{Metric IV: Gel Condition}
The sensor's soft gel is covered by a textured fabric skin for protection. Experimentally, this significantly increases the resilience to wear. However, the reflective paint layer of the gel may still wear out after use.
Since the specular paint acts as a reflection surface, the regions of the gel with damaged paint do not respond to contact signal and are seen as black pixels, which we call \emph{dead pixels}. 

\begin{figure}[h]
\centering
  \includegraphics[width=\linewidth]{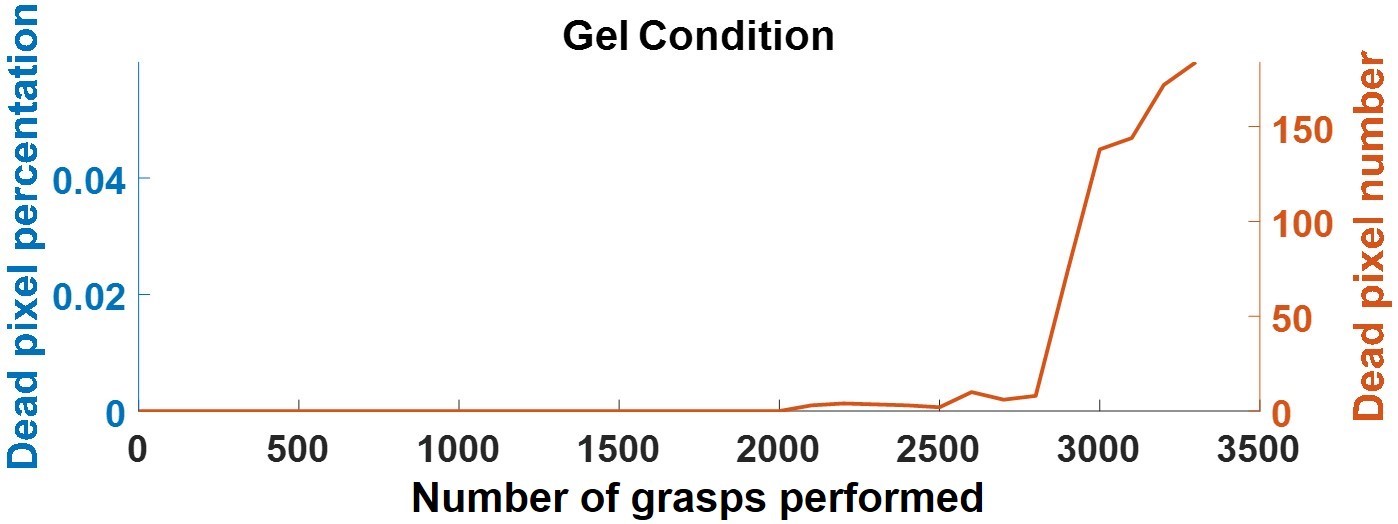}
  \caption{Evolution of the gel condition.}
  \label{fig:gel_condition}
  \vspace{-3mm}
\end{figure}

We define the gel condition as the percentage of dead pixels in the image.  \figref{fig:gel_condition} shows the evolution of the number of dead pixels over the course of 3000 grasps. Only a small amount of pixels (less than 0.06\%, around 170 pixels) are damaged, highlighted with red circles in \figref{fig:step2} (b3).
%
Sparse dead pixels can be ignored or fixed with interpolation, but a large number of clustered dead pixels can be solved only by replacing the gel. 

\section{Conclusions and Future Work}
\label{sec:conclusion}
In this paper, we present a compact integration of a visual-tactile sensor in a robotic phalange. Our design features: a gel covered with a textured fabric skin that improves durability and contact signal strength; a compact integration of the GelSight sensor optics; and an improved illumination over a larger tactile area. 
Despite the improved wear resistance, the sensor still ages over use. We propose four metrics to track this aging process and create a calibration framework to regularize sensor output over time. We show that, while the sensor degrades minimally over the course of several thousand grasps, the digital calibration procedure is able to condition the sensor output to improve its usable life-span.

\myparagraph{Sensor Functionality.}
\change{The sensor outputs images of tactile imprints that encode shape and texture of the object at contact. For example, contact geometry in pixel space could be used in combination with knowledge of grasping force and gel material properties to infer 3D local object geometry. If markers are placed on the gel surface, marker flow can be used to estimate object hardness~\cite{GelSightIROS16} or shear forces~\cite{GelSightShear}.}
These quantities, as well as the sensor's calibrated image output, can be used directly in model-based or learning-based approaches to robot grasping and manipulation. This information could be used to track object pose, inform a data-driven classifier to predict grasp stability, or as real-time observations in a closed-loop regrasp policy~\cite{Frank_Maria_regrasp}.
%
%

\myparagraph{Applications in robotic dexterity.}
\change{We anticipate that GelSlim's unique form factor will facilitate the use of these sensing modalities in a wide variety of applications -- especially in cluttered scenarios where visual feedback is lacking, where access is limited, or where difficult to observe contact forces play a key role. We are especially interested in using real-time contact geometry and force information to monitor and control tasks that require in-hand dexterity and reactivity such as picking a tool in a functional grasp and then using it or grasping a nut and screwing it on a bolt. Ultimately, these contact-rich tasks can only be robustly tackled with tight integration of sensing and control. While the presented solution is just one path forward, we beilive that high-resolution tactile sensors hold particular promise.}




\bibliographystyle{IEEEtranN} 
{\footnotesize \bibliography{main}} 

\end{document}